# Detecting Levels of Depression in Text Based on Metrics

Ashwath Kumar Salimath
Trinity College Dublin
salimata@tcd.ie

Robin K Thomas
Trinity College Dublin
rothomas@tcd.ie

Sethuram Ramalinga Reddy
Trinity College Dublin
ramalins@tcd.ie

Yuhao Qiao
Trinity College Dublin
qiaoy@tcd.ie

## 1 Abstract

Depression is one of the most common and a major concern for society. Proper monitoring using devices that can aid in its detection could be helpful to prevent it all together. The Distress Analysis Interview Corpus (DAIC) is used to build a metric-based depression detection. We have designed a metric to describe the level of depression using negative sentences and classify the participant accordingly. The score generated from the algorithm is then levelled up to denote the intensity of depression. The results show that measuring depression is very complex to using text alone as other factors are not taken into consideration. Further, In the paper, the limitations of measuring depression using text are described, and future suggestions are made.

## 2 Introduction

A problem of undiagnosed people suffering from depression persists in most of the world. A usual screening process is an online questionnaire, and it is a self-selective process. If a person completes a survey, then that person already suspects a problem. Depression is a major mental health disorder that is rapidly affecting lives worldwide. Depression impacts not only emotional but also a physical and psychological state of the person. Its symptoms include lack of interest in daily activities, feeling low, anxiety, frustration, loss of weight and even feeling of self-hatred.

The data for the analysis was obtained from DAIC-WOZ [1]. The dataset consists of conversations with 189 participants. We are concentrating on the replies by the participant because we measure the depression levels of a participant. The idea is to detect the level of depression using a custom-metric. By focussing on the Participants data, the depression metric can be generalised to a larger population. The total emotion considering all the replies given by a participant is generated to find the depression. The process involves calculating sentiment polarity on each word and adding weights based on the intensity and summing up to get the emotion score.

Our research question is to design a metric to describe the level of depression and classify the participant accordingly. In future, the focus would be to apply the model to a broader area, such as social media, and to achieve a scale-based and automated measure of depression. The research paper is structured in four parts: Literature Review, Methodology and Implementation, Results and Limitations, Conclusion and Future Work.

# 3 Literature Review

In the below section we describe the psychological condition that we study, namely depression. We refer to previous works on depression levels, the metric to measure the intensity and the various methods to measure them using semantic analytics. Despite many years of intense research, the measure for depression levels and the intensity has not been measured properly. One of the reason might be due to the difficulty in using the medium for measurement like speech or text, each of which has its challenges. The other reason might be since proper quantized data is not available to be used for proper research and the available data is at times inconsistent. Another challenge is the representation of emotion, how the behaviour is observed and how to score the emotional intensity of a person. Despite these challenges, sentiment analysis has been very popular to measure the depression levels and has provided many methods which are widely used to estimate sentiments, i.e., measure the level of depression from the user point of view.

Ann Devitt et al. [2] provides a basic understanding of sentiment polarity identification. They use a cohesion-based text representation method used to compare news stories and group the texts based on polarity. Here polarity direction and intensity is used as evaluation metric relative to the human decisions. A reference to the work done by Pang and Viswanathan was mentioned wherein they worked on identifying polarity based on the absence of or preens of particular texts. Devitts work also relies on polarity measurement based on the lexicon of pre-tagged positive and negative terms which are used as relative positive and negative emotion descriptor. The intensity was measured based on average human ratings for specific text, and as such, this is not always a linear relationship between average ranking scale and the real intensity displayed through the text. Also, the problem of over-reliance on the hand-coded lexicon had its cons as they will be inconsistent and prone to errors. So the use of multiple lexicons from different unique sources was suggested so that together using multiple lexicons would increase the base polarity values and a better match with human notions. Rajdeep Singh et.al.[3], uses a Lexicon based sentiment analyser to determine the popularity of an online post. This method uses polarity to index words and the identification of adjectives, adverbs and verbs to assign the final polarity to a sentence to determine the sentiment of the author. For this purpose, a polarity of 4,3,2 was assigned to adjectives, adverbs and verbs respectively. Using the general statistics associated with the comment, it was classified as positive, negative or neutral and then based on the above-mentioned parts of speech, the final polarity of the comment is calculated. The main issue was the use of slang words which could be used in a positive way as well as a negative way, but it depends on the actual way the person uses. It provided a better view of understanding and perceiving different sentences and words. Manfred Klenner et al. [4], take the above method one step further with the use of an Affect Lexicon that provided the prior polarity of the words. Then a chunker is used to determine words that are relevant to the uses in compositional phrase level polarity determination. Here all the words in the sentence are not used, and specific phrases are used to determine the final polarity. The main problem identifying the words were regarding neutral words and their compositions. The composed phrase might create a positive or a negative term which might be overlooked by the method. The sentiment analysis using lexicon sentiment composition provides a better way to understand the polarity and sentiment evaluation given prior polarity.

Yair Neuman et al. [5], provides a method to screen depression-related words proactive metaphorical screening. This method extracts metaphorical relations in which depression or relevant subsets of this term is used and extracts the conceptual domains relating to it. A lexicon is generated using these results, which is used to evaluate the level of depression in texts automatically or whether the whole text deals with the depression domain. Although this was a unique way the domain level analysis would prove costly, and the apt measure of depression may not be achieved. The level of depression is also not achieved using this method as it was based on the frequency of words in the texts, which can at times provide inconsistent results. ToMeisha Edwards et al. [6], describe the correlation between the use of first person singular pronouns and the depression levels of a person. The use of first-person pronouns has been previously linked as a marker for depression by Ireland and Mehl [7] in 2014. ToMeisha and others have found the idea that depressed people are in a state of more self-reflection and tend to focus on themselves and talk about themselves frequently even during a conversation. Using this method, singular first-person pronouns were evaluated in text transcripts, and it was found that there was indeed a correlation between more frequent use of first of singular first-person pronouns when people are depressed. The main analysis to note was that this effect was common among all demographics including age and gender. But the usage of this method could only tell to the extent that the person was feeling depressed, i.e. the sentence would be a negative polarity sentence, but the level or intensity of the polarity was not discussed.

Jonathan Gratch et. al.[8], of the Distress Analysis Interview Corpus conducted several clinical interviews designed to support the psychological condition. These interviews were conducted by humans as well as by an agent named Ellie. This data has been transcribed and is being used in this research. The transcribed data has provided a list of words with the term frequency which tells us which words the users used more. They also used multiple clinical methods like the use of sensors, respiration of the patients and Electro Cardio Gram to determine the emotional state of the person. Additionally, the automated agents were able to generate logs of the user's speech and provide real-time recognition of the user's voice pattern and the use of repetitive words. It was found that the people displayed emotions more intensely when interacting with a computer compared to a human. This transcript generated through these interviews will be used in the current research. Although this has been used and investigated in the past, using a new method to identify the words, it would be possible to avoid the major hindrances that do not allow for the measurement of the emotional intensity of the user. The data generated here contains clinical interviews designed to support the diagnosis of psychological distress conditions such as anxiety, depression, and post-traumatic stress disorder.

## 4 Methodology

1. Dataset Description

    The dataset was obtained from DAIC-WOZ Database. This dataset is produced from the conversation between the participants (Happy and sad participants) and human controlled chat software. The dataset consists of interviews with 189 participants. This dataset has their response in the form of voice recordings, facial expressions, and the text. We consider only text dataset which is named as TRANSCRIPT.CSV. This data consists of start and stop time of the sentence spoken, the speaker and the value that was spoken by the speaker. We are concentrating on the replies given by the participant because we are analysing the data, in the context of depression level of the participant and not the replies given to the questions asked by the software. By this way, we can generalise the depression metric to the large population.

2. Data Preparation

    This step involves transforming the data to extract only the useful information and remove the unwanted or non-informative texts. This includes following steps:

    (a) To delete the replies or sentences framed by the software Ellie.
    (b) To separate the start and stop time from each participants reply.
    (c) To tokenise the replies into words by using the whitespace as the delimiter.
    (d) To remove stop words from the corpus by making use of the dictionary of stop words collected from the Internet.
    (e) To use filters to remove the markers which indicate words like mmm, hmm etc., Since we cannot derive any useful information from them.
    (f) To remove the words that are presented in the angular brackets like ¡laugh¿ that indicates an action done by the participant within the replies.

(g) To reduce the complexity, we can delete the reply that has got only actions like <sigh>, <p>etc.

3. Algorithm

    The Depression level of the people can be derived from the conversation by using the following approaches:

    (a) Intensity-Based Approach
        i. First, we start analysing from the lowest granular level which is from the words, and we will check whether that word is positive or negative.
        ii. If they are either positive or negative, then we have to give weight to each word based on the intensity of their emotion.
        iii. We also need to find the words that are related to the depression and give it more weight than the other sad words. For, E.g. the word suicide must be given more weight in this case.
        iv. Now for each line, we will measure the total emotion by summing up the magnitude along with the respective sentiment polarity
        v. In this way, we can measure the emotion of all the lines, and by averaging them, we can get the overall measure of the sentiment in the form of some percentage. For, E.g. If the result is 20%, then the person is happy, or else if the result is 80%, then the person is sad.

    (b) Duration-Based Approach
        i. We calculate the overall duration of the conversation from the starting time of the first reply and the stopping time of the last reply.
        ii. Then we can divide the whole duration into equal splits of time intervals, and we can find the average of emotion as calculated in the previous approach for each interval.
        iii. If we plot this average over the intervals, then we can also see his mood swing, and if larger the number of intervals dominates in the depression, then he requires immediate attention by the doctors.
        iv. Then if we find the total average for all these intervals, we can get the overall mood. This is a way to evaluate whether the value obtained is similar to that of the value obtained in earlier approach.
        v. We can also evaluate our measure of overall emotion with the established result given in the dataset that the particular conversation is a happy or a sad conversation.

## 5 Implementation

We have considered Intensity-Based Approach for the Implementation. We have used Google Cloud Natural API to process the Transcripts to detect the depression level. The API inspects the given text and identifies the prevailing emotional opinion within the text, especially to determine a writer's attitude as positive, negative, or neutral. It attempts to determine the overall attitude (positive or negative) expressed within the text. The sentiment is represented by numerical score and magnitude values.

We designed a scale to capture the level of depression based on the participant's transcript. The Scale is developed based on three variables, i.e. X (Number of Negative Sentences in a Transcript/Total Number of Sentences in a Transcript), Y (Mean of the Score of all Negative Sentences in a Transcript), Z (Mean of the Magnitude of all Negative Sentences in a Transcript). The sentences which have scored between -1.0 and -0.25 are considered. Our Focus is only on Negative Sentences as we are detecting depression which is a negative state.

The number of Negative Sentences in a Transcript, Y and Z values are the outputs given by Google Cloud Natural Language API [9]. The scale is the sum of three variables, i.e. X, Y/2 and Z/4. Y refers to the score of the sentiment which ranges between -1.0 and -0.25 (positive) and corresponds to the overall emotional leaning of the text. Z refers to the Magnitude which indicates the strength of negative emotion within the given text, between 0.0 and +inf. Unlike Y, Z is not normalized; each expression of emotion within the text contributes to the text's magnitude (so longer text blocks may have greater magnitudes). To reduce the effect of an unnormalized score, i.e. Z in our scale, we have transformed Z into Z/4.

# 6 Results

We have used dev_split_Depression_AVEC2017.CSV file as a Reference to evaluate the performance of our algorithm. This file comprises of participant IDs, PHQ8 Binary labels, PHQ8 Scores, and participant gender, and single responses for every question of the PHQ8 questionnaire for the official development split. A PHQ8 score is the sum of the eight responses given by a Participant in the questionnaire. A score of 10 or greater is considered major depression, 20 or more is severe major depression. For Example, Transcript Number 439 has a score of 1 as per PHQ8, So, we consider this Participant as Normal.

For Example, Transcript Number 439 when fed into the Algorithm, It gave a score of 29.455, where it consists of X, Y/2 and Z/4 components. The Final Score is then Further classified on Four Levels, i.e. Score 0 - 25 (Happy Participant), 26-50 (Low Depressed), 51-75 (Medium Depressed) and 76-100 (High Depressed). So, the Participant Number 439 will belong to Low Depressed category. The output from both PHQ8 Questionnaire and our Algorithm for the sample transcripts is described in the table below.

| Transcript Number | Algorithm Score | PHQ8 Score |
|---|---|---|
| 302 | 20.2 | 2 |
| 346 | 36 | 23 |
| 367 | 28 | 19 |
| 382 | 12 | 0 |
| 439 | 29 | 1 |
| 440 | 19 | 19 |
| 482 | 25 | 1 |

Table 1: Algorithm and PHQ8 Scores

# 7 Limitations

The output of the Depression Metric Algorithm was analyzed, and we found that there is no way to evaluate our Algorithm's Performance. The scores generated by our Algorithm considers only the sentences with Negative Scores as we are focused on finding the level of depression. We have only PHQ8 Questionnaire Scores to be used as a reference to assess our Algorithm's performance for some sample Transcripts.

For the evaluation, The Algorithm scores cannot be compared with PHQ8 Questionnaire scores due to the Following reasons:

1. The Person who is in depression might not express sincerely in an Interview, which makes Algorithm's output Questionable as we are using Transcript Text for analysis. But, The Person might have a score on PHQ8 Questionnaire above 10 describing depression.

2. PHQ8 Questionnaire Score and the Algorithm Score are both calculated in different ways. PHQ8 Questionnaire Score is calculated based on the responses given by a participant, whereas, Algo-rithm Score is calculated based on Sentiment Analysis as per the Google Cloud Natural Language API.

## 8 Conclusion

This paper proposed a method for the sentiment analysis of the conversation with 189 participants. This dataset gives us voice recordings, facial expressions, and the text. We dropped the sentence replied by "Ellie". Then removed the stopwords in the corpus and used filters to remove the flags for words like MMM, HMM, <laugh>as we cannot get any useful information from them. As mentioned above, we could choose two algorithms to measure the level of depression. We could check whether one word is positive or sad, and find the words associated with depression and make them more critical than other sad words. We could measure the emotions of each line in the conversation and average the emotions by taking a percentage of them. Or we can calculate the total duration of the conversations from the start of the first reply to the stop of the last reply. Then we can divide the entire length into equal splits of time intervals, and we can find the average of sentiment calculated in previous intervals in each interval.

We used Google Cloud Natural API to process the Transcripts to detect the depression level. The sentiment is represented by numerical score and magnitude values. We designed a scale to capture the level of depression based on the participant's transcript. The Scale is developed based on three variables. We have used dev_split_Depression_AVEC2017.CSV file as a Reference to evaluate the performance of our algorithm. The Final Score is classified into Four Levels, i.e. Score 0 - 25 (Happy Participant), 26-50 (Low Depressed), 51-75 (Medium Depressed) and 76-100 (High Depressed). In future work, we need to find solutions to deal with the limitations mentioned above, then finally apply the model to a broader area, such as social media, and to achieve a scale-based and automated measure of depression. We expect that the model will identify the individual's need for help and the possibility of self-harm.

## References

[1] Available at: *http://dcapswoz.ict.usc.edu/*. Accessed April 4, 2018.

[2] Devitt, Ann, and Khurshid Ahmad. "Sentiment polarity identification in financial news: A cohesion-based approach." In *Proceedings of the 45th annual meeting of the association of computational linguistics*, pp. 984-991. 2007.

[3] Singh, Rajdeep, Roshan Bagla, and Harkiran Kaur. "Text analytics of web posts' comments using sentiment analysis." In *Computing and Communication (IEMCON), 2015 International Conference and Workshop on*, pp. 1-5. IEEE, 2015.

[4] Klenner, Manfred, Stefanos Petrakis, and Angela Fahrni. "A tool for polarity classification of human affect from panel group texts." In *Affective Computing and Intelligent Interaction and Workshops, 2009. ACII 2009. 3rd International Conference on*, pp. 1-6. IEEE, 2009.

[5] Neuman, Yair, Yohai Cohen, Dan Assaf, and Gabbi Kedma. "Proactive screening for depression through metaphorical and automatic text analysis." *Artificial intelligence in medicine* 56, no. 1 (2012): 19-25.

[6] To'Meisha Edwards, Nicholas S. Holtzman. "A meta-analysis of correlations between depression and first person singular pronoun use." (2017).

[7] Ireland, Molly E., and Matthias R. Mehl. "Natural language use as a marker." *The Oxford handbook of language and social psychology* (2014): 201-237.

[8] Gratch, Jonathan, Ron Artstein, Gale M. Lucas, Giota Stratou, Stefan Scherer, Angela Nazarian, Rachel Wood et al. "The Distress Analysis Interview Corpus of human and computer interviews." In *LREC*, pp. 3123-3128. 2014.

[9] Available at: *https://cloud.google.com/natural-language/docs/basics*. Accessed April 4, 2018.